\documentclass[runningheads]{llncs}
\usepackage[T1]{fontenc}
\usepackage{graphicx}

\usepackage{subcaption}
\usepackage{dblfloatfix} 
\usepackage{multirow}
\usepackage{tabularx,booktabs,hhline} 
\usepackage{enumitem}
\usepackage[linesnumbered,ruled,vlined]{algorithm2e}
\SetKwInput{KwInput}{Input}                
\SetKwInput{KwOutput}{Output} 
\usepackage{soul}
\begin{document}
\title{XAI for time-series classification leveraging image highlight methods}
%
%
\author{Georgios Makridis\inst{1}\orcidID{0000-0002-6165-7239} \and
Georgios Fatouros\inst{1,2}\orcidID{0000-0001-6843-089X} \and
Vasileios Koukos\inst{1}\orcidID{}\and
Dimitrios Kotios\inst{1}\orcidID{}\and
Dimosthenis Kyriazis \inst{1}\orcidID{0000-0001-7019-7214}\and
Ioannis Soldatos \inst{3}\orcidID{ht0000-0002-6668-3911}}
\authorrunning{pre-print - }
%
\institute{University of Piraeus, Pireas 185 34, Greece \\  \and
Innov-Acts Ltd, City Business Center Office SF09, 27 Michalakopoulou St. 1075, Nicosia, CY
\email{info@innov-acts.com} \and
Netcompany-Intrasoft S.A, 2b, rue Nicolas Bové L-1253 Luxembourg}

\maketitle

\begin{abstract}
Although much work has been done on explainability in the computer vision and natural language processing (NLP) fields, there is still much work to be done to explain methods applied to time series as time series by nature can not be understood at first sight. In this paper, we present a Deep Neural Network (DNN) in a teacher-student architecture (distillation model) that offers interpretability in time-series classification tasks. The explainability of our approach is based on transforming the time series to 2D plots and applying image highlight methods (such as LIME and GradCam), making the predictions interpretable. At the same time, the proposed approach offers increased accuracy competing with the baseline model with the trade-off of increasing the training time. 
\end{abstract}


\section{Introduction}\label{sec1}

\begin{figure*}[h]
\centering
\includegraphics[width=0.9\textwidth]{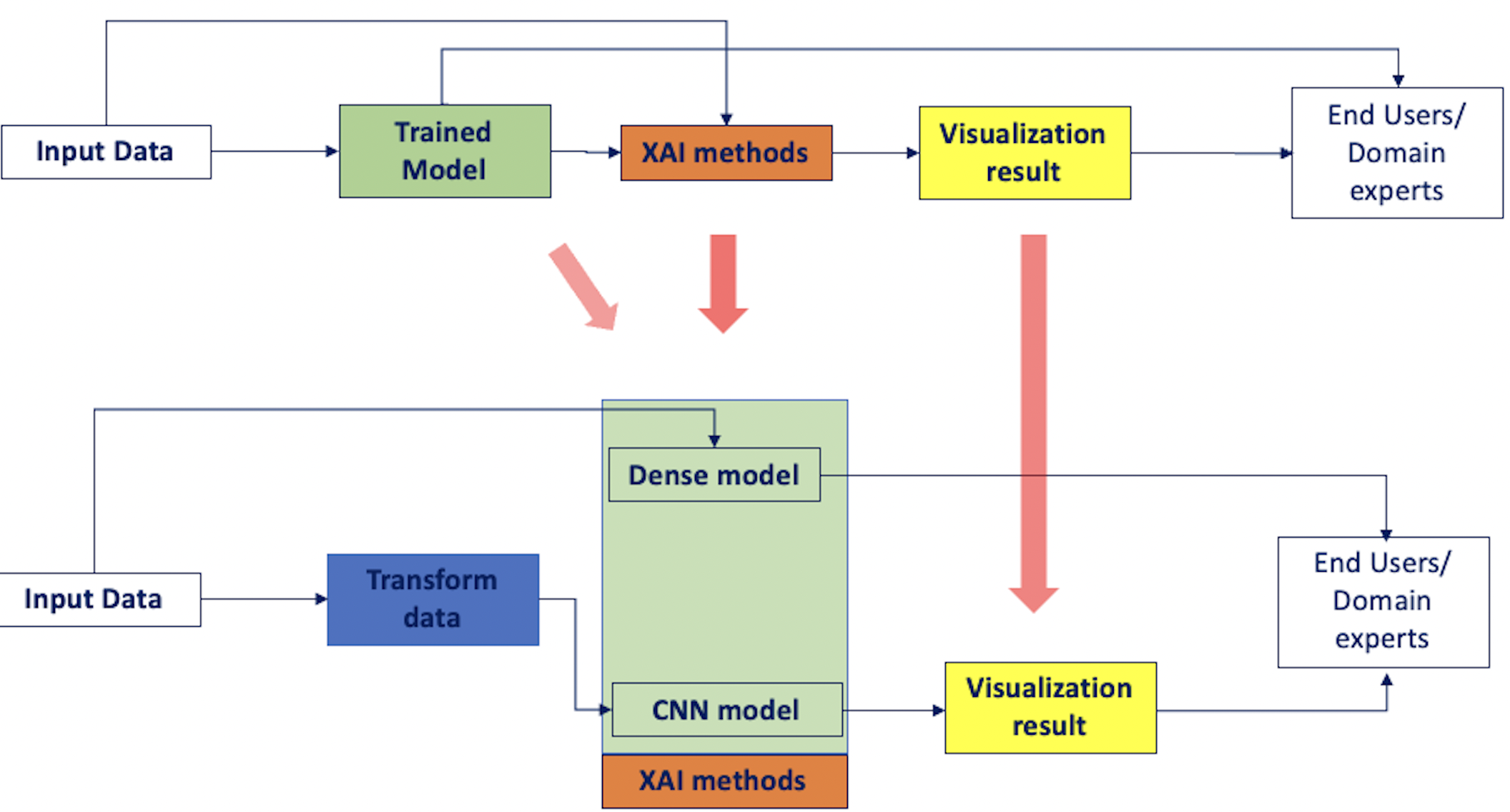}
\caption{The proposed high-level architecture. The above pipeline represents the "traditional" way of applying XAI methods, while the lower one depicts the proposed model. }
\label{fig:high_level_arch}
\end{figure*}

\par The current day and age, also known as the Digital or Information Age, is characterized by complex computing systems which generate enormous amounts of data daily. The digital transformation of industrial environments leads to the fourth industrial revolution -Industry4.0 \cite{makridis2020predictive}, with Artificial Intelligence (AI) being the key facilitator of the Industry4.0 era by enabling innovative tools and processes \cite{soldatos2021trusted}. At the same time, there has been a growing interest in eXplainable Artificial Intelligence (XAI) towards human-understandable explanations for the predictions and decisions made by machine learning models. 

\par The notion of explaining and expressing a Machine Learning (ML) model is called interpretability or explainability \cite{choo2018visual}. This need for interpretability mainly exists in Deep Neural Networks (DNN), which are defined by large levels of complexity, thus appearing to be "black boxes" \cite{zahavy2016graying}. To address this issue, researchers have proposed various methods for explaining the predictions made by deep learning models for image classification tasks, known as XAI for images \cite{zeiler2014visualizing}. These methods include visualizing the features learned by the model, feature attribution techniques, and model distillation \cite{shrikumar2017learning}. For example, visualizing the features learned by the model can provide insights into which parts of the input image the model uses to make its prediction, while feature attribution techniques can highlight the regions of the input image that are most important for making a prediction. Model distillation, on the other hand, attempts to convert a complex model into a simpler, more interpretable one that still preserves the accuracy of the original model. A knowledge distillation system consists of three principal components: the knowledge, the distillation algorithm, and the teacher-student architecture \cite{gou2021knowledge}.  Model distillation is a technique where a smaller, simpler model is trained to mimic the predictions of a larger, more complex model. This allows for the smaller model to be more easily interpretable and can help to provide insight into the workings of the larger model. For example, in \cite{chiu20201st} authors proposed a method called "DeepTaylor" that uses a Taylor series approximation to distill a deep neural network into a linear model that is more interpretable.  

\begin{figure*}
\centering
\includegraphics[width=\textwidth]{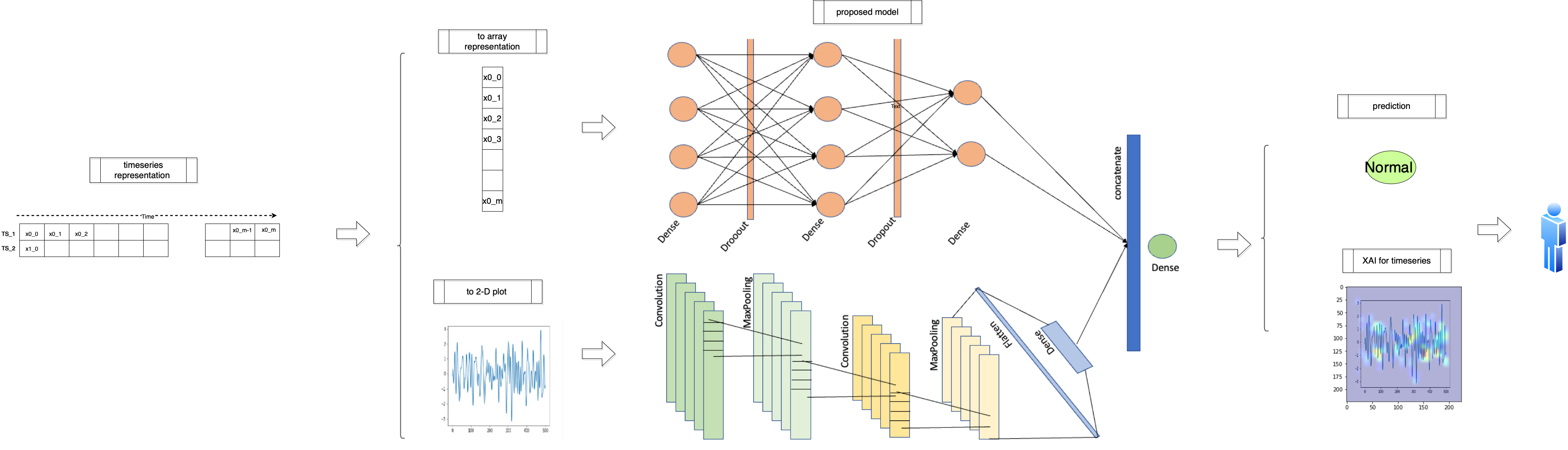}
\caption{Schematic explanation of the data aspect to the end-to-end approach including the main steps to the process, from raw data to the user.}
\label{fig:data_pipeline}
\end{figure*}

\par Although much work has been done on explainability in the computer vision and Natural Language Processing (NLP) fields, there is still much work to be done to explain methods applied to time series. This might be caused by the nature of time series, which we can not understand at first sight. Indeed, when a human looks at a photo or reads a text, he intuitively and instinctively understands the underlying information in the data \cite{rojat2021explainable}. Although temporal data is ubiquitous in nature, through all forms of sound, humans are not used to representing this temporal data in the form of a signal that varies as a function of time. We need expert knowledge or additional methods to leverage the underlying information present in the data \cite{lim2021time}. This problem might explain why a lot of existing methods focus on highlighting the parts of the input responsible for a prediction for computer vision tasks (i.e., image classification). Despite this drawback, timeseries approach play a key role in industry 4.0 so does XAI for these cases (e.g., XAI for timeseries can enhance transparency in Industry 4.0 by providing clear and interpretable explanations for how the predictive maintenance tool works, 
or for identifying patterns and anomalies in time series data, providing early detection of potential issues, and insights into how to mitigate risks).

\par This paper addresses the emerging challenge of explainability of the "black-box" DNN models concerning time series as input data by introducing a teacher-student (distillation model) approach to facilitate user explainability and reasoning.  In this context, the operational objective is twofold:
\begin{enumerate}[label=(\roman*)]
    \item Time series categorization based on efficient classification of each time series, providing a sound approach for time series data along with unstructured image data (i.e., 2D plot) can be classified using AI-based methods and represented in the time domain as time-series enables the exploitation of the image XAI methods. 
    \item Systematically providing visualization of XAI by applying various XAI methods on the CNN part of the proposed model. 
\end{enumerate}

\par While the scientific contributions that support and extend the proposed solution can be briefly summarised in terms of added value, as the proposed approach enables:  
\begin{enumerate}[label=(\roman*)]
    \item Combination of the strengths of both networks to achieve better performance. 
    \item Handling of both time series as arrays and as images, which provides more flexibility for data inputs and potentially enhances the interpretability of the model.
    \item Explainability by visualizing the important parts of the plot for time series, which is an open research field with limited state-of-the-art solutions.
\end{enumerate}

\par In this context, the proposed approach presents a novel approach to XAI for time series data, in which image highlight techniques are used to visualize and interpret the predictions of a time series model as depicted in Fig. \ref{fig:high_level_arch}. While a more detailed view of the proposed approach is provided in Fig. \ref{fig:data_pipeline}, which depicts the data handling pipeline, in an approach, to apply both the time series classification and XAI methods. In the first step, the raw data are pre-processed and fed into the distillation model. Then the model is trained and finally, the output comprises the predicted "class" and the heatmap of the time-series representation to be utilized for explainability purposes.

\par The remainder of the paper is organized as follows: Section 2 presents related work done in the areas of study of this paper, while Section 3 delivers an overview of the proposed methodological approach, introduces the overall architecture, and offers details regarding the datasets used and how these are utilized within the models. Section 4 dives deeper into the results of the conducted research and the implemented algorithms, with the performance of the proposed mechanisms being depicted in the results and evaluation section. Section 5 concludes with recommendations for future research and the potential of the current study. 

\section{Related Work}

\par Time series classification is a challenging task, where DNNs have been shown to achieve state-of-the-art results in various fields such as finance \cite{fatouros2022deepvar}, and industry \cite{makridis2020enhanced}. However, the internal workings of these models are often difficult to interpret. To address this issue, researchers have proposed various methods for explaining the predictions made by deep learning models for time series classification, known as XAI for time series. These methods include attention mechanisms \cite{mitchell2019model} and visualization techniques \cite{dahl2018private}. For example, attention mechanisms allow the model to focus on specific parts of the input time series that are important for making a prediction, while visualization techniques can provide a graphical representation of the input time series and the model's decision-making process. 

\par Attention mechanisms allow the model to focus on specific parts of the input time series, providing insight into which regions of the input are most important for making a prediction. For example, in \cite{bach2015pixel} authors proposed an attention-based LSTM model that learns to weigh different parts of the input time series when making a prediction.

\par Furthermore there are post-hoc methods that approach the behavior of a model by exporting relationships between feature values and predictions, in this case, feature lags of time series also play a role. While the ante-hoc methods incorporate the explanation in the structure of the model, which is therefore already explainable at the end of the training phase.
The first category includes variants of the widely used LIME, k-LIME \cite{hall2017machine}, DLIME \cite{zafar2019dlime}, LIMEtree \cite{sokol2020limetree}, and SHAP (e.g., TimeSHAP \cite{bento2021timeshap}), as well as Anchors \cite{ribeiro2018anchors}, or LoRE \cite{guidotti2018local}. Surrogate models are built for each prediction sample in most of these techniques, learning the behavior of the reference model in the particular instance of interest by adding perturbations (or masking) to the feature vector variables. The numerous feature disturbance techniques for assessing the contribution of features to the projected value when they are deleted or covered are nearly a distinct research topic in this context. These XAI models may be used for DNN since they are unaffected by the underlying machine-learning model. The CAM, ConvTimeNet \cite{kashiparekh2019convtimenet}, which belongs to this group but intervenes in the model structure, is particularly interesting. Apart from these, it is worth mentioning approaches such as RETAIN \cite{choi2016retain} (Reverse Time Attention) with application to Electronic Health Records (EHR) data. RETAIN achieves high accuracy while remaining clinically interpretable and is based on a two-level neural attention model that detects previous visits. Finally, NBEATS \cite{oreshkin2019n} focuses on an architecture based on residual back-and-forth connections and a very deep stack of fully connected layers. The architecture has a number of desirable properties, as it is interpretable and applicable without modification to a wide range of target areas.

\par One limitation of XAI models for time series data is that the outputs and explanations they provide may be difficult for end users to understand.  The models' complexity and the explanations' technical nature may make it challenging for non-experts to interpret the results and take appropriate actions. Additionally, the effectiveness of XAI models can depend on the end user's level of domain knowledge and familiarity with the data. To address these limitations, it is important to design XAI models with the end user in mind and provide explanations that are tailored to their needs and level of expertise. This involve proper visualizations to help users explore and understand the model's predictions, as well as providing contextual information about them following the "human-centricity" perspective that is the core value behind the evolution of manufacturing towards Industry 5.0 \cite{rovzanec2022human}. 

\section{Method}

\subsection{Proposed Model Architecture}

\par
\begin{figure*}
\centering
\includegraphics[width=\textwidth]{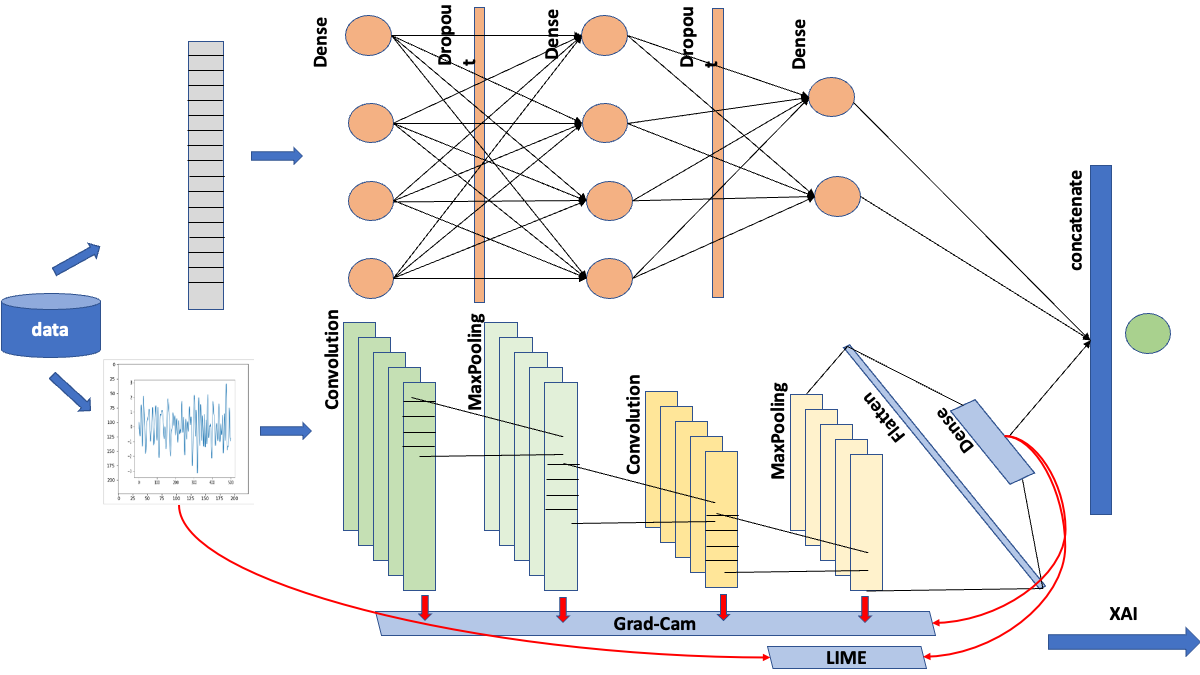}
\caption{The architecture of the proposed model, where the Dense model represents the "teacher" while the CNN represents the "explainable"-"student". The data are fed as an array to the Dense network while the CNN as 2D plots.}
\label{fig:model_arch}
\end{figure*}

 The main research challenges addressed in this paper enable the accurate classification of time series and the generation of interpretable XAI visualizations. These two topics are interconnected as the XAI model is based on the classification results. To facilitate the usage of a DNN model by making it explainable, we introduce a form of distillation model (i.e., teacher-student architecture) where the teacher is the complex DNN model that yields state-of-the-art performance while the student model is a CNN model that can be interpretable by human end users when LIME and GradCam methods are applied to it. Fig. \ref{fig:model_arch} depicts the architecture of the proposed model, where the Dense model represents the "teacher" while the CNN represents the "explainable"-"student".

\par Regarding the DNN  model, its layers can be summarized as follows:

\begin{itemize}
    \item Input layer: This layer will take an image as input.
    \item Rescaling layer.
    \item Convolutional layers: to extract features from the input image. The number of filters and their size can be chosen based on the complexity of the data. In our experiments, we used 32 filters of size 3x3, and a stride of 1.
    \item Pooling layers:  to reduce the spatial dimensions of the feature maps and reduce the computational requirements of the network. A common choice is to use max pooling with a pool size of 2x2 and a stride of 2.
    \item Flattening layer: to convert the high-dimensional feature maps into a 1D vector that can be input into a fully connected layer.
    \item Dense layer: One or multiple dense (fully connected) layers can be added to make predictions based on the extracted features. The number of nodes in each dense layer can be chosen based on the complexity of the data and the desired performance of the model.
    \item Output layer: this can be a dense layer with a single node and a sigmoid activation function for binary classification or with as many nodes as the number of classes in your data for multi-class classification.
\end{itemize}

\subsection{Datasets}

In the frame of our research, the utilized datasets have been retrieved by \cite{dau2019ucr} composed of labeled time series. Specifically, we used 2 of the available datasets with more than 1000 training and test samples: 
\begin{itemize}
    \item Wafer: This dataset was formatted by R. Olszewski as part of his thesis Generalized feature extraction for structural pattern recognition in time-series data at Carnegie Mellon University, 2001. Wafer data relates to semiconductor microelectronics fabrication. A collection of inline process control measurements recorded from various sensors during the processing of silicon wafers for semiconductor fabrication constitute the wafer database; each data set in the wafer database contains the measurements recorded by one sensor during the processing of one wafer by one tool. The two classes are normal and abnormal. There is a large class imbalance between normal and abnormal (10.7\% of the train are abnormal, 12.1\% of the test). The best performance in terms of Accuracy till now is 99,8\%  by using the ST algorithm \cite{lines2012shapelet}.
    \item FordA: This data was originally used in a competition at the IEEE World Congress on Computational Intelligence, 2008. The classification problem is to diagnose whether a certain symptom exists or does not exist in an automotive subsystem. Each case consists of 500 measurements of engine noise and a classification. There are two separate problems: For FordA the train and test data set were collected in typical operating conditions, with minimal noise contamination. The best performance in terms of Accuracy till now is 96,54\%  by using the ST algorithm. 
    
\end{itemize}

\subsection{Time-series to Image explainability} 

\par In order to make the time series interpretable we chose the most comprehensive representation of a time series (i.e., a 2D plot with the time axis on the x-axis and the value of the time series on the y-axis.) We first plot the time series as an image and then input the image into the CNN for classification. In all of our experiments, we plotted the time series as line-plot figures with (224,244,1) dimensions. 

Once we have plotted the time series as an image, we use them as inputs into a CNN for classification. The CNN will process the image and use its learned features to make a prediction about the time series.

\begin{figure}
\centering
\includegraphics[width=0.8\linewidth]{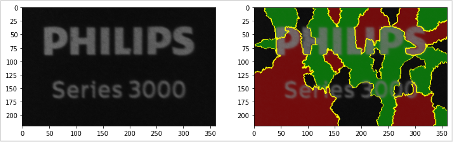}
\caption{Example of LIME important super-pixels}
\label{fig:lime_philips}
\end{figure}

\par When dealing with image data, feature importance can be translated into the importance of each pixel to the output forecast. The latter can also be visualized into a heatmap where the importance of each feature (pixel) can be displayed with different colors and can work as an excellent explanation for the human operator. In our approach, we have chosen 2 notable methods of XAI on image data as the more appropriate for time series plots. The LIME model is a model-agnostic method based on the work of \cite{ribeiro2016should}. While, the Gradient-weighted Class Activation Mapping (Grad-Cam) model \cite{selvaraju2017grad} was used as one of the SotA methods for interpreting the top features (parts of the image) concerning the "Label" when the underlying model is a CNN. To identify the abstracted notion of a given image, Grad-Cam utilizes gradients to generate localization maps and highlight essential parts of the image. Despite the high level of complexity, Grad-Cam is able to provide intuitive outputs, thus improving the model's accuracy and flexibility. 

In Fig. \ref{fig:lime_philips}, we see the output of an XAI model that has been applied to a plot of a time series based on the LIME method in the study \cite{Makridis2022}, where we can see the green and red areas of the image. This means that the ML/DL model classifies the underlying image to a specific category because of these parts of super-pixels; the size of super-pixels colored in green are the ones that increase the probability of our image belonging to the predicted class, while the super-pixels colored in red are the ones that decrease the likelihood. LIME is a technique for explaining the predictions of any machine learning model by approximating it locally with an interpretable model. It does this by perturbing the input data and creating a new, simpler model that is fit only to the perturbed data. This process is repeated many times, each time with a different set of perturbations, and the results are aggregated to create a global explanation of the model's prediction. In the case of image data, LIME can be used to explain the prediction of a CNN or other image classification model by creating a heatmap that shows which parts of the image are most important for the model's prediction. The heatmap is created by perturbing the pixels of the image and creating a new, simpler model that is fit only to the perturbed data. This process is repeated for different parts of the image, and the results are aggregated to create the final heatmap. 

\begin{figure}
\centering
\includegraphics[width=0.8\linewidth]{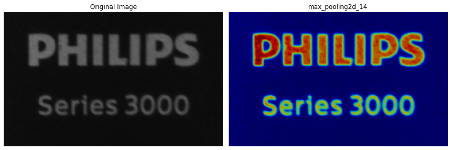}
\caption{GradCam heatmap examples in the last layer of the CNN model}
\label{fig:gradcam_philips}
\end{figure}

\par An example of applying the Grad-Cam method is depicted in Fig.\ref{fig:gradcam_philips}.  We can see the original image and the explanation of grad-cam methods where there are highlighted parts of the image that the CNN model is focusing on in each layer. This is a technique for visualizing the regions in an image that are most important for a convolutional neural network (CNN) to classify an image correctly. It does this by overlaying a heatmap on top of the original image, where the intensity of the color in the heatmap indicates how important that region is for the classification. To create the heatmap, Grad-CAM first computes the gradient of the output of the CNN with respect to the feature maps of a convolutional layer. It then weighs these gradients by the importance of the feature maps to the class prediction and produces a weighted combination of the feature maps. Finally, it resizes the resulting map to the size of the input image and overlays it on top of the original image to produce the heatmap. Overall, the heatmap produced by Grad-CAM can be used to understand which regions of an image are most important for a CNN's prediction and to identify potential areas of the image that may be causing the CNN to make incorrect predictions.

\section{Evaluation Results}
\label{results}

\subsection{Experimental Setup}

\par We evaluate the performance of our model on a classification task, using a train, validation, and test dataset split. The task involves classifying a set of time series into one of two classes. Our distillation model consists of a teacher network, which is a Dense network, and a student network, which is a CNN. The goal of the distillation is to transfer the knowledge from the teacher network to the student network, resulting in an explainable and more efficient model with similar or better performance. Given that Model distillation also has its own limitations, one of them is that it may not always be possible to distill a complex model into a simpler, more interpretable model. To this end, we provide the results of the teacher-student network, the teacher-only and the student-only results. 

\par We use the training dataset to train the models, and a validation dataset to tune the hyperparameters of the network, such as the learning rate and the number of layers. We perform early stopping if the validation accuracy does not improve for a certain number of epochs. Finally, we evaluate the performance of the teacher and student networks on the test dataset.

\par We compare the performance of the teacher network, the student network, and our distillation model. We report the accuracy, precision, recall, F1-score for each model. Additionally, we showcase the interpretability of the student network using LIME and GradCAM, which highlight the important parts of the input images that contribute to the classification decision.

\par We use the TensorFlow deep learning framework to implement the models and train them in an environment with 20 CPU cores of 2.3GHz, and 64GB of RAM. The hyperparameters are chosen based on a grid search over a range of values, and the best set of hyperparameters are chosen based on the validation accuracy. The models are trained for a maximum of 300 epochs, with early stopping if the validation accuracy does not improve for 20 epochs. The code and the trained models are publicly available for reproducibility. Moreover, we make a qualitative evaluation regarding the interpretability of the XAI result of our approach. Finally, we present a more fine-grained view of the performance of the examined method in terms of time. The results are the average outcomes of ten independently seeded runs for each measurement.

\subsection{Experimental Results}

\par For both the FordA and Wafer datasets, the Dense network achieved the lowest precision and F1-score for class 1 compared to the other models. This suggests that the Dense network may not be as effective at identifying the minority class in imbalanced datasets.
The CNN and proposed models achieved perfect precision and F1-score for class 0 on both datasets, indicating that they were able to effectively classify the majority class. The proposed model also achieved the highest precision and F1-score for class 1 on both datasets, suggesting that it may be better suited to handling imbalanced datasets.
The overall accuracy of the models varied, with the Dense network achieving the lowest accuracy on the FordA dataset, and the CNN achieving the lowest accuracy on the Wafer dataset. The proposed model achieved the highest accuracy on both datasets

\begin{table}[h]
\centering
\caption{Results of the classification tasks with different models}
\label{tab:combined_results}
\begin{tabular}{cccc}
\toprule
\textbf{Model} & \textbf{(FordA) Dense} & \textbf{(FordA) CNN} & \textbf{(FordA) Proposed} \\
\midrule
Precision (Class 0) & 0.86 & 0.70 & 0.86 \\
Precision (Class 1) & 0.82 & 0.75 & 0.84 \\
F1-Score (Class 0) & 0.84 & 0.73 & 0.86 \\
Accuracy & 0.84 & 0.73 & 0.85 \\
\midrule
\textbf{Model} & \textbf{(Wafer) Dense} & \textbf{(Wafer) CNN} & \textbf{(Wafer) Proposed} \\
\midrule
Precision (Class 0) & 1.0 & 1.0 & 1.0 \\
Precision (Class 1) & 0.98 & 1.0 & 0.99 \\
F1-Score (Class 0) & 1.0 & 1.0 & 1.0 \\
Accuracy & 1.0 & 1.0 & 1.0 \\
\bottomrule
\end{tabular}
\end{table}

\begin{table}[]
\centering
\caption{Execution time of each model for the training task as well as how many epochs took each model to converge. The RAM column refers to the MB needed for loading the training datasets to the RAM}
\label{tab:my-table}
\resizebox{\linewidth}{!}{%
\begin{tabular}{|l|c|c|c|}
\hline
 & \textit{\textbf{trianing time (sec)}} & \textit{\textbf{early stopping (epochs)}} & \textit{\textbf{train-set RAM (MB)}} \\ \hline
\textbf{Dense (FordA)}       & 31   & 67  & 1.16    \\ \hline
\textbf{CNN (FordA)}         & 643  & 76  & 1148.44 \\ \hline
\textbf{Dense + CNN (FordA)} & 5106 & 102 & 1149.6  \\ \hline
\textbf{Dense (Wafer)}       & 186  & 300 & 13.74   \\ \hline
\textbf{CNN (Wafer)}         & 3539 & 300 & 4135.52 \\ \hline
\textbf{Dense + CNN (Wafer)} & 3048 & 257 & 4149.26 \\ \hline
\end{tabular}%
}
\end{table}

\par Given that, time performance can impact the scalability and cost-effectiveness of AI systems, as slower models may require more computational resources or increased processing time, leading to increased costs. Thus, optimizing the performance of AI models in terms of time is a crucial aspect of AI development and deployment. To this end, in Table \ref{tab:my-table} we present the execution time of each model for the training task as well as how many epochs took each model to converge. We should highlight that the configurations of the models were identified as mentioned in the previous subsection. It should also be mentioned that concerning the inference time (i.e., the time taken by the model to produce a prediction for a single input) there is no significant discrepancy.  

\par
\begin{figure}
\centering
\includegraphics[width=0.7\linewidth]{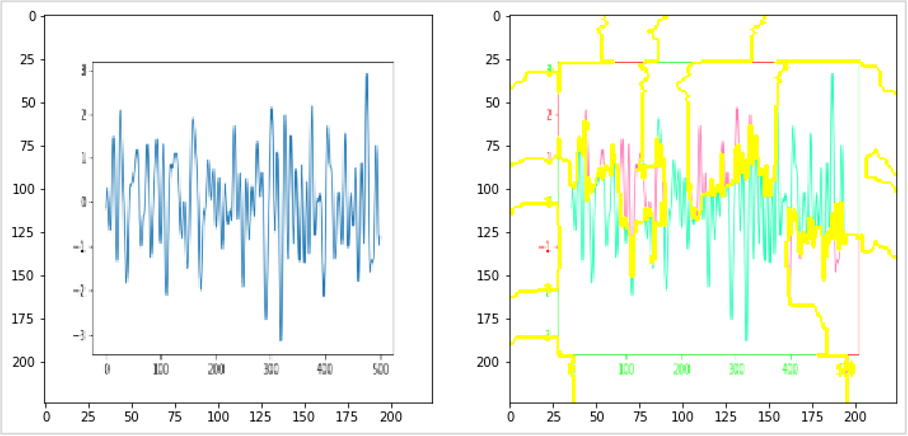}
\caption{Example of LIME important superpixels where we can notice the red areas that seem not relevant to the predicted class }
\label{fig:lime_time}
\end{figure}

\begin{figure}[!t]
\centering
\begin{subfigure}[b]{0.7\linewidth}
\includegraphics[width=\linewidth]{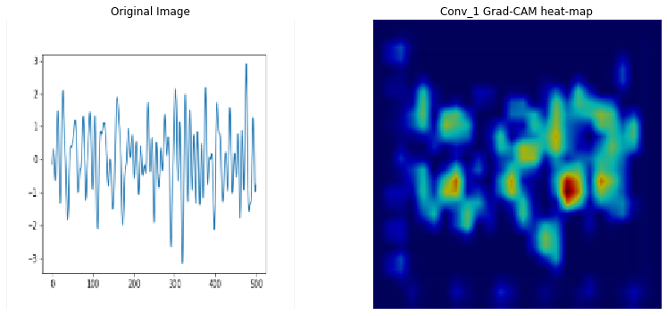}
\caption{Example of Grad-Cam heatmap and the original image where we can notice the areas of the image that the model focused}
\end{subfigure}
\begin{subfigure}[b]{0.7\linewidth}
\centering
\includegraphics[width=0.5\linewidth]{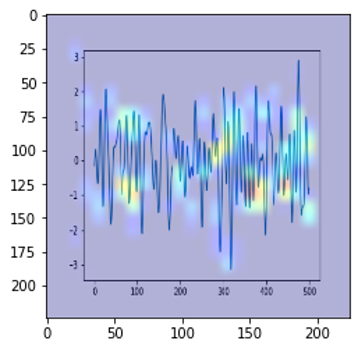}
\caption{Example of GradCam heatmap and the overlap of heatmap to the original image where we can notice the areas of the image that the model focused}
\end{subfigure}
\caption{Grad-Cam example in the last convolution layer of the CNN}
\label{fig:gradcam_time}
\end{figure}

 Moving to the "explainability" part of the model, the model has integrated the process to provide LIME and Grad-Cam heatmaps helping the interpretation for a non-data scientist user. LIME helps to understand why a machine learning model made a certain prediction by looking at the most important parts of the input data that the model considered when making the prediction. While Grad-CAM generates a heatmap that shows which areas of an image were most important for a machine learning model to make a prediction. This helps us to see which parts of the image the model focused on to make its prediction. In Fig.\ref{fig:lime_time} and Fig.\ref{fig:gradcam_time} we can see some indicative examples of the respective XAI visualizations.

\section{Conclusion}

\par In conclusion, this paper examines an XAI-enhanced DNN for addressing the problem of interpretability of DNN models when it comes to time series classification tasks. Specifically, this paper presents a novel distillation model for time-series classification that leverages image highlight methods for XAI. The model takes in both the time series data as an array and a 2D plot, providing a unique approach to time-series classification. The student model is specifically designed for XAI images, demonstrating the potential of this method for improving interpretability in time-series classification tasks. The results of the experiments conducted show that the proposed model outperforms existing methods and has the potential to be applied to a wide range of time-series classification problems.   A teacher-student architecture was developed where the student part has as inputs the 2D plot of the time series where LIME and GradCam model is applied offering XAI capabilities. Regarding the future steps, we have 2 main objectives to make the model more human-centered by applying a model to translate the XAI visualizations to text, of course, we plan to apply the model to time series forecasting tasks while at the same time examining visual analytics approach, where the human would inspect the figures and provide some kind of feedback.

\par As future work we plan to develop a similar approach for time series forecasting tasks.

\subsubsection{Acknowledgements}
The research leading to the results presented in this paper has received funding from the Europeans Union’s funded Projects MobiSpaces under grant agreement no 101070279 and STAR under grant agreement no 956573.

\bibliographystyle{splncs04}
\bibliography{refs}

\begin{thebibliography}{10}
\providecommand{\url}[1]{\texttt{#1}}
\providecommand{\urlprefix}{URL }
\providecommand{\doi}[1]{https://doi.org/#1}

\bibitem{bach2015pixel}
Bach, S., Binder, A., Montavon, G., Klauschen, F., M{\"u}ller, K.R., Samek, W.:
  On pixel-wise explanations for non-linear classifier decisions by layer-wise
  relevance propagation. PloS one  \textbf{10}(7),  e0130140 (2015)

\bibitem{bento2021timeshap}
Bento, J., Saleiro, P., Cruz, A.F., Figueiredo, M.A., Bizarro, P.: Timeshap:
  Explaining recurrent models through sequence perturbations. In: Proceedings
  of the 27th ACM SIGKDD Conference on Knowledge Discovery \& Data Mining. pp.
  2565--2573 (2021)

\bibitem{chiu20201st}
Chiu, M.T., Xu, X., Wang, K., Hobbs, J., Hovakimyan, N., Huang, T.S., Shi, H.:
  The 1st agriculture-vision challenge: Methods and results. In: Proceedings of
  the IEEE/CVF Conference on Computer Vision and Pattern Recognition Workshops.
  pp. 48--49 (2020)

\bibitem{choi2016retain}
Choi, E., Bahadori, M.T., Sun, J., Kulas, J., Schuetz, A., Stewart, W.: Retain:
  An interpretable predictive model for healthcare using reverse time attention
  mechanism. Advances in neural information processing systems  \textbf{29}
  (2016)

\bibitem{choo2018visual}
Choo, J., Liu, S.: Visual analytics for explainable deep learning. IEEE
  computer graphics and applications  \textbf{38}(4),  84--92 (2018)

\bibitem{dahl2018private}
Dahl, M., Mancuso, J., Dupis, Y., Decoste, B., Giraud, M., Livingstone, I.,
  Patriquin, J., Uhma, G.: Private machine learning in tensorflow using secure
  computation. arXiv preprint arXiv:1810.08130  (2018)

\bibitem{dau2019ucr}
Dau, H.A., Bagnall, A., Kamgar, K., Yeh, C.C.M., Zhu, Y., Gharghabi, S.,
  Ratanamahatana, C.A., Keogh, E.: The ucr time series archive. IEEE/CAA
  Journal of Automatica Sinica  \textbf{6}(6),  1293--1305 (2019)

\bibitem{fatouros2022deepvar}
Fatouros, G., Makridis, G., Kotios, D., Soldatos, J., Filippakis, M., Kyriazis,
  D.: Deepvar: a framework for portfolio risk assessment leveraging
  probabilistic deep neural networks. Digital Finance pp. 1--28 (2022)

\bibitem{gou2021knowledge}
Gou, J., Yu, B., Maybank, S.J., Tao, D.: Knowledge distillation: A survey.
  International Journal of Computer Vision  \textbf{129},  1789--1819 (2021)

\bibitem{guidotti2018local}
Guidotti, R., Monreale, A., Ruggieri, S., Pedreschi, D., Turini, F., Giannotti,
  F.: Local rule-based explanations of black box decision systems. arXiv
  preprint arXiv:1805.10820  (2018)

\bibitem{hall2017machine}
Hall, P., Gill, N., Kurka, M., Phan, W.: Machine learning interpretability with
  h2o driverless ai. H2O. ai  (2017)

\bibitem{kashiparekh2019convtimenet}
Kashiparekh, K., Narwariya, J., Malhotra, P., Vig, L., Shroff, G.: Convtimenet:
  A pre-trained deep convolutional neural network for time series
  classification. In: 2019 International Joint Conference on Neural Networks
  (IJCNN). pp.~1--8. IEEE (2019)

\bibitem{lim2021time}
Lim, B., Zohren, S.: Time-series forecasting with deep learning: a survey.
  Philosophical Transactions of the Royal Society A  \textbf{379}(2194),
  20200209 (2021)

\bibitem{lines2012shapelet}
Lines, J., Davis, L.M., Hills, J., Bagnall, A.: A shapelet transform for time
  series classification. In: Proceedings of the 18th ACM SIGKDD international
  conference on Knowledge discovery and data mining. pp. 289--297 (2012)

\bibitem{makridis2020predictive}
Makridis, G., Kyriazis, D., Plitsos, S.: Predictive maintenance leveraging
  machine learning for time-series forecasting in the maritime industry. In:
  2020 IEEE 23rd International Conference on Intelligent Transportation Systems
  (ITSC). pp.~1--8. IEEE (2020)

\bibitem{makridis2020enhanced}
Makridis, G., Mavrepis, P., Kyriazis, D., Polychronou, I., Kaloudis, S.:
  Enhanced food safety through deep learning for food recalls prediction. In:
  International Conference on Discovery Science. pp. 566--580. Springer (2020)

\bibitem{Makridis2022}
Makridis, G., Theodoropoulos, S., Dardanis, D., Makridis, I., Separdani, M.M.,
  Fatouros, G., Kyriazis, D., Koulouris, P.: Xai enhancing cyber defence
  against adversarial attacks in industrial applications. In: 2022 IEEE 5th
  International Conference on Image Processing Applications and Systems (IPAS).
  pp. xx--xx (2022). \doi{10.xxxx/xxxx.xxxxxxx}

\bibitem{mitchell2019model}
Mitchell, M., Wu, S., Zaldivar, A., Barnes, P., Vasserman, L., Hutchinson, B.,
  Spitzer, E., Raji, I.D., Gebru, T.: Model cards for model reporting. In:
  Proceedings of the conference on fairness, accountability, and transparency.
  pp. 220--229 (2019)

\bibitem{oreshkin2019n}
Oreshkin, B.N., Carpov, D., Chapados, N., Bengio, Y.: N-beats: Neural basis
  expansion analysis for interpretable time series forecasting. arXiv preprint
  arXiv:1905.10437  (2019)

\bibitem{ribeiro2016should}
Ribeiro, M.T., Singh, S., Guestrin, C.: " why should i trust you?" explaining
  the predictions of any classifier. In: Proceedings of the 22nd ACM SIGKDD
  international conference on knowledge discovery and data mining. pp.
  1135--1144 (2016)

\bibitem{ribeiro2018anchors}
Ribeiro, M.T., Singh, S., Guestrin, C.: Anchors: High-precision model-agnostic
  explanations. In: Proceedings of the AAAI conference on artificial
  intelligence. vol.~32 (2018)

\bibitem{rojat2021explainable}
Rojat, T., Puget, R., Filliat, D., Del~Ser, J., Gelin, R.,
  D{\'\i}az-Rodr{\'\i}guez, N.: Explainable artificial intelligence (xai) on
  timeseries data: A survey. arXiv preprint arXiv:2104.00950  (2021)

\bibitem{rovzanec2022human}
Ro{\v{z}}anec, J.M., Novalija, I., Zajec, P., Kenda, K., Tavakoli~Ghinani, H.,
  Suh, S., Veliou, E., Papamartzivanos, D., Giannetsos, T., Menesidou, S.A.,
  et~al.: Human-centric artificial intelligence architecture for industry 5.0
  applications. International Journal of Production Research pp. 1--26 (2022)

\bibitem{selvaraju2017grad}
Selvaraju, R.R., Cogswell, M., Das, A., Vedantam, R., Parikh, D., Batra, D.:
  Grad-cam: Visual explanations from deep networks via gradient-based
  localization. In: Proceedings of the IEEE international conference on
  computer vision. pp. 618--626 (2017)

\bibitem{shrikumar2017learning}
Shrikumar, A., Greenside, P., Kundaje, A.: Learning important features through
  propagating activation differences. In: International conference on machine
  learning. pp. 3145--3153. PMLR (2017)

\bibitem{sokol2020limetree}
Sokol, K., Flach, P.: Limetree: Interactively customisable explanations based
  on local surrogate multi-output regression trees. arXiv preprint
  arXiv:2005.01427  (2020)

\bibitem{soldatos2021trusted}
Soldatos, J., Kyriazis, D.: Trusted Artificial Intelligence in Manufacturing: A
  Review of the Emerging Wave of Ethical and Human Centric AI Technologies for
  Smart Production. Now Publishers (2021)

\bibitem{zafar2019dlime}
Zafar, M.R., Khan, N.M.: Dlime: A deterministic local interpretable
  model-agnostic explanations approach for computer-aided diagnosis systems.
  arXiv preprint arXiv:1906.10263  (2019)

\bibitem{zahavy2016graying}
Zahavy, T., Ben-Zrihem, N., Mannor, S.: Graying the black box: Understanding
  dqns. In: International conference on machine learning. pp. 1899--1908. PMLR
  (2016)

\bibitem{zeiler2014visualizing}
Zeiler, M.D., Fergus, R.: Visualizing and understanding convolutional networks.
  In: Computer Vision--ECCV 2014: 13th European Conference, Zurich,
  Switzerland, September 6-12, 2014, Proceedings, Part I 13. pp. 818--833.
  Springer (2014)

\end{thebibliography}



\end{document}